\documentclass[letterpaper]{article} 
\usepackage{aaai2026}  
\usepackage{times}  
\usepackage{helvet}  
\usepackage{courier}  
\usepackage[hyphens]{url}  
\usepackage{graphicx} 
\usepackage{natbib}  
\usepackage{caption} 
\usepackage{algorithm}
\usepackage{algorithmic}

\usepackage{newfloat}
\usepackage{listings}
\DeclareCaptionStyle{ruled}{labelfont=normalfont,labelsep=colon,strut=off} 
\floatstyle{ruled}
\newfloat{listing}{tb}{lst}{}
\floatname{listing}{Listing}
\title{Beyond the Black Box: Demystifying Multi-Turn LLM Reasoning with VISTA}
\author {
    Yiran Zhang\textsuperscript{\rm 1},
    Mingyang Lin\textsuperscript{\rm 2},
    Mark Dras\textsuperscript{\rm 1},
    Usman Naseem\textsuperscript{\rm 1}
}
\affiliations {
    \textsuperscript{\rm 1}Macquarie University\\
    \textsuperscript{\rm 2}Independent Researcher\\
    \{yiran.zhang, mark.dras, usman.naseem\}@mq.edu.au
}

\begin{document}

\maketitle

\begin{abstract}
Recent research has increasingly focused on the reasoning capabilities of Large Language Models (LLMs) in multi-turn interactions, as these scenarios more closely mirror real-world problem-solving. However, analyzing the intricate reasoning processes within these interactions presents a significant challenge due to complex contextual dependencies and a lack of specialized visualization tools, leading to a high cognitive load for researchers. To address this gap, we present \textbf{VISTA}, an web-based \textbf{V}isual \textbf{I}nteractive \textbf{S}ystem for \textbf{T}extual \textbf{A}nalytics in multi-turn reasoning tasks. VISTA allows users to visualize the influence of context on model decisions and interactively modify conversation histories to conduct ``what-if" analyses across different models. Furthermore, the platform can automatically parse a session and generate a reasoning dependency tree, offering a transparent view of the model's step-by-step logical path. By providing a unified and interactive framework, VISTA significantly reduces the complexity of analyzing reasoning chains, thereby facilitating a deeper understanding of the capabilities and limitations of current LLMs. The platform is open-source and supports easy integration of custom benchmarks and local models.
\end{abstract}

\begin{links}
    \link{Code}{https://github.com/grantzyr/vista-platform}
\end{links}

\section{Introduction}

Large Language Models (LLMs) show strong performance across natural language tasks such as generation, summarization, and question answering. As their abilities grow, research focus has shifted from single-turn benchmarks to more complex multi-turn interactions~\citep{zhang2025turnbenchmsbenchmarkevaluatingmultiturn, golde2025mastermindevalsimplescalablereasoning, light2023avalonbenchevaluatingllmsplaying, ruoss2025lmactbenchmarkincontextimitation,tang2025dsgbenchdiversestrategicgame,laban2025llmslostmultiturnconversation}. These scenarios require models to track conversational state, interpret evolving context, and perform multi-step reasoning, reflecting more realistic human problem-solving.

Despite rising interest, evaluating reasoning in multi-turn dialogue remains difficult. Model decisions depend on information spread across long histories, and tracing contextual cues behind specific outputs or errors often requires time-consuming manual inspection. The lack of interactive visualization tools limits systematic diagnosis, making it hard to attribute failures or test counterfactual cases.

To address this gap, we introduce VISTA, an interactive web-based visual analytics platform for analyzing multi-turn reasoning in LLMs. VISTA turns conversation logs into transparent, interactive workspaces, helping researchers visualize context, trace logic, and compare behaviors across sessions. Our work makes the following key contributions:

\begin{itemize}
    \item \textbf{A Novel Visual Analytics Platform}: VISTA provides an intuitive user interface to visualize the entire multi-turn reasoning and decision-making process. This design enables researchers to easily trace and analyze the relationship between the prior context and model output.   
    
    \item \textbf{Interactive Counterfactual Analysis}: The platform supports dynamic interaction by allowing users to modify any part of the dialogue history, and re-run inference with the same or different models. This facilitates powerful counterfactual analyses through the comparison of parallel sessions. The system also generates reasoning trees to illustrate step-by-step logic.
    
    \item \textbf{An Extensible Architecture}: We provide a flexible framework with a decoupled back-end that allows for integration of custom benchmarks. It employs a standardized API for model management, enabling straightforward connection and analysis of various models, including those hosted locally by the user.
\end{itemize}

\begin{figure*}[!t]
  \centering
  \includegraphics[width=\textwidth]{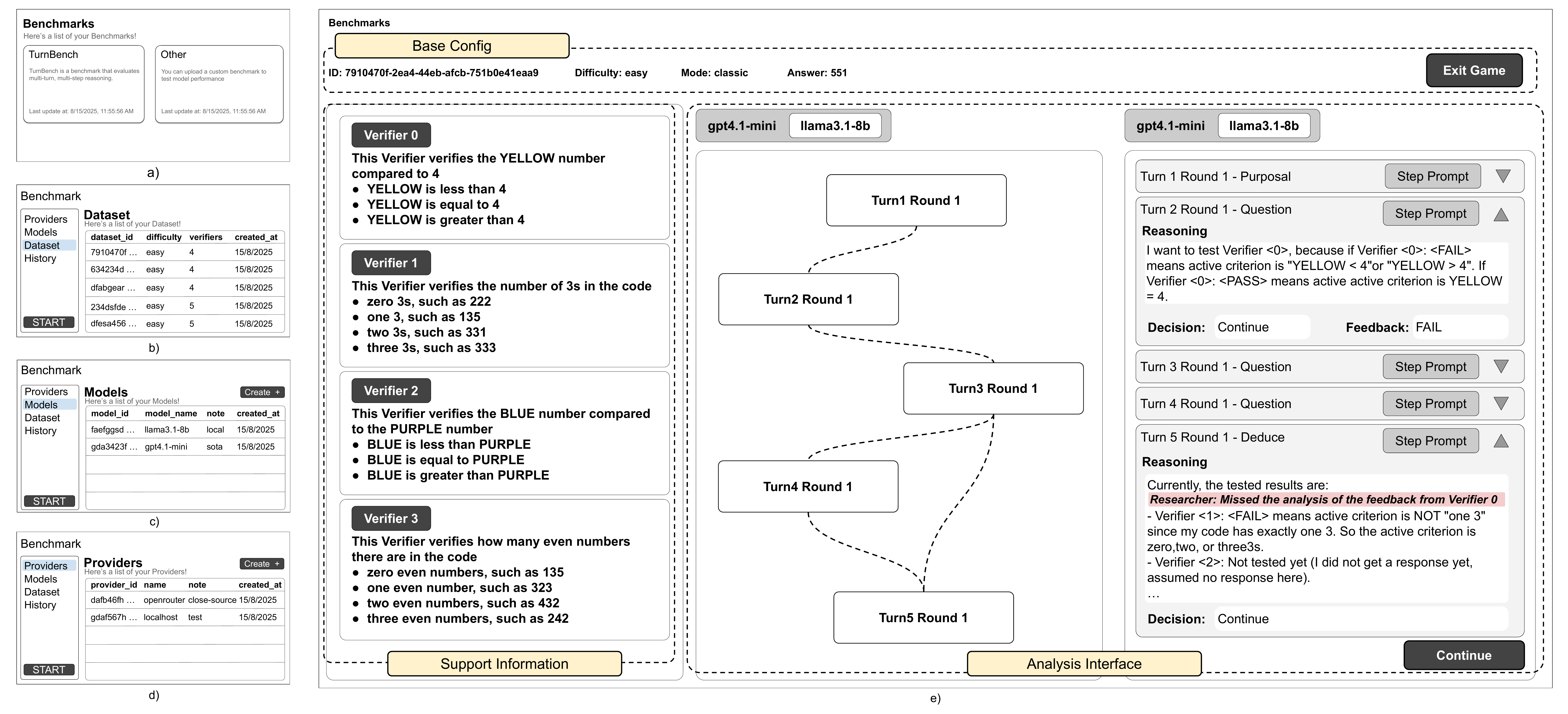}
  \caption{Overview of the five main system pages: (a) Benchmark management; (b) Dataset management; (c) Model management; (d) Provider management; (e) Benchmarking page, showing dataset settings at the top, support information (e.g., verifiers for multi-step reasoning) on the left, and an analysis interface on the right for visualizing reasoning steps, model decisions, and editing prompts or outputs for new sessions.}
  \label{fig:overall_figure}
\end{figure*}

\section{VISTA Platform Framework}

Our platform is designed to analyze the reasoning processes of LLMs in multi-turn interactions, an area that remains difficult to study \cite{huang-chang-2023-towards}. While prior tools like ReasonGraph \cite{li-etal-2025-reasongraph} focus on single-turn visualization, our system provides the first end-to-end tool for multi-turn scenarios.

The system adopts a client-server architecture. The front-end built with React, supports interactive visualization. The backend implemented in Python with FastAPI, manages logic through modular components for benchmarks, model settings, providers, reasoning analysis, sessions, and data setups (Figure \ref{fig:overall_figure}). Sessions and interactions are stored in PostgreSQL, enabling persistent and resumable analysis. This modular design allows easy extension, e.g., adding new models or benchmarks through standardized templates.

The platform supports three main capabilities:

\paragraph{Unified Model and Benchmark Management} VISTA provides a centralized UI to manage model configurations and sources. Users can easily add, remove, or modify models and their parameters. A standardized API facilitates the integration of locally hosted models, enabling direct comparison between commercial APIs (e.g., GPT-4), open-source models, and proprietary fine-tuned models. Similarly, new benchmarks can be integrated by adhering to a provided template interface, as demonstrated by the pre-loaded TurnBench benchmark \cite{zhang2025turnbenchmsbenchmarkevaluatingmultiturn}.

\paragraph{Interactive Counterfactual Analysis} A key feature of our platform is the support for dynamic interaction. Researchers can modify any message within a dialogue history and either apply the change to the current session or spawn a new, parallel session driven by the same or a different model. This facilitates powerful counterfactual analyses, allowing for direct observation of how contextual changes or model choices impact subsequent reasoning paths.

\paragraph{Automated Reasoning Explanation} To elucidate the model's internal logic, VISTA can leverage a powerful LLM to automatically parse a conversation and generate a Reasoning Dependency Tree. This feature renders a structured, graphical representation of the model's step-by-step inference process, making intricate logical chains transparent and easier to scrutinize for errors or biases.

\section{Case Study}

To demonstrate VISTA’s utility, we conduct a case study with the TurnBench benchmark~\citep{zhang2025turnbenchmsbenchmarkevaluatingmultiturn}, which evaluates multi-turn, multi-step reasoning by requiring models to iteratively use verifier feedback across rounds to eliminate inconsistent options. For a detailed description of the benchmark, see \citet{zhang2025turnbenchmsbenchmarkevaluatingmultiturn}.

In a typical workflow, a researcher selects a model and runs a TurnBench session step by step. When an illogical move occurs, the Reasoning Dependency Tree helps reveal errors, such as reliance on outdated or irrelevant information. For example, in Figure \ref{fig:overall_figure} (right panel), VISTA highlights a mistake where the model disregards Verifier 0’s feedback and makes a logically inconsistent deduction. This makes the failure immediately visible and guides the researcher to edit the dialogue history and launch a parallel session to test the hypothesis. If the revised session corrects the mistake, the counterfactual analysis confirms the cause.

This case illustrates how VISTA turns LLM debugging into a concrete, interactive process, enabling precise tracing and validation of reasoning errors.

\section{Conclusion}

We presented VISTA, an open-source visual analytics platform for in-depth analysis of multi-turn reasoning in Large Language Models. By combining interactive visualization with counterfactual exploration and automated reasoning explanations, VISTA turns complex reasoning failure debugging from an ad-hoc activity into a structured, evidence-driven process. With its extensible design supporting both custom models and benchmarks, VISTA lowers the barrier for fine-grained diagnostics of LLM behavior and helps accelerate the community’s understanding of their strengths and limitations.

\bibliography{aaai2026}

\end{document}